\documentclass[sigconf]{acmart}
\AtBeginDocument{%
  \providecommand\BibTeX{{%
    \normalfont B\kern-0.5em{\scshape i\kern-0.25em b}\kern-0.8em\TeX}}}

\newcommand{\diff}{\mathit{diff}}

\begin{document}

\title{A Novel Multiagent Flexibility Aggregation Framework}


\author{Stavros Orfanoudakis}
\affiliation{%
  \institution{Technical University of Crete}
  \city{Chania}
  \country{Greece}
}
\email{sorfanoudakis@tuc.gr}

\author{Georgios Chalkiadakis}
\affiliation{%
  \institution{Technical University of Crete}
  \city{Chania}
  \country{Greece}
}
\email{gehalk@intelligence.tuc.gr}

\begin{abstract}
The increasing number of Distributed Energy Resources (DERs) in the emerging Smart Grid, has created an imminent need for intelligent multiagent frameworks able to utilize these assets efficiently. In this paper, we propose a novel DER aggregation framework, encompassing a multiagent architecture and various types of mechanisms for the effective management and efficient integration of DERs in the Grid. One critical component of our architecture is the Local Flexibility Estimators (LFEs) agents, which are key for offloading the Aggregator from serious or resource-intensive responsibilities---such as addressing privacy concerns and predicting the accuracy of DER statements regarding their offered demand response services. The proposed  framework allows the formation of efficient LFE cooperatives. To this end, we developed and deployed a variety of cooperative member selection mechanisms, including {\em (a)} scoring rules, and {\em (b)} (deep) reinforcement learning. 
We use data from the well-known PowerTAC simulator to systematically evaluate our framework. Our experiments verify its effectiveness for incorporating heterogeneous DERs into the Grid in an efficient manner.
In particular, when using the well-known probabilistic prediction accuracy-incentivizing CRPS scoring rule as a selection mechanism, our framework results in increased average payments for participants, when compared with traditional commercial aggregators.
\end{abstract}





\maketitle

\section{Introduction}
The depletion of fossil fuels has created an imminent need to deploy even more renewable energy power generators~\cite{Berrill2016EnvironmentalIO}. However, the state of the current energy grid makes it hard to efficiently optimize the performance of intermittent assets~\cite{Adefarati2016IntegrationOR}, such as solar panels and wind turbines; thus, the need for a ``smarter'' electricity grid has been created~\cite{Ramchurn2012PuttingT,Judge2022OverviewOS}. The Smart Grid, with its bidirectional electricity and information flow, is envisaged to deliver electrical power in very resourceful ways, and successfully exploit all the Distributed Energy Resources (DERs) that are continuously emerging. DERs are the various electricity supply or demand assets that are spread across the Grid; and which, however small, when combined, can enhance the Grid's ability to seamlessly provide power, even if it largely originates from intermittent renewable energy sources.

Furthermore, recent developments regarding environmental policies and the emergence of a multitude of (distributed) energy markets, have turned the attention of the electricity stakeholders to the research on DERs' flexibility~\cite{Lampropoulos2019AFF}. In the literature, \textit{flexibility} is defined as the elasticity property of the DERs that can provide ancillary services to support the stability of the grid~\cite{Sawall2018FlexibilityDF}. 
Essentially, flexibility corresponds to the DERs' ability to either offer produced/stored energy or  consumption reduction services to the Grid.
Thus, many works are studying various ways to estimate the flexibility of DER assets~\cite{Iraklis2021FlexibilityFA,DeZotti2019ConsumersFE}; while the use of {\em aggregators}~\cite{Gkatzikis2013TheRO,Okur2021AggregatorsBM,tesi_ogzeokur} is one of the most important mechanisms that were created to utilize the flexibility of the DERs in the Smart Grid.

An aggregator is a mediator between DERs and the energy markets~\cite{Gkatzikis2013TheRO}, with the mission to trade the flexibility obtained from the DERs by participating in the markets on behalf of the DERs' owners~\cite{tesi_ogzeokur}. Generally, aggregators offer stability guarantees to the Grid by offering flexible loads. Currently, the existing legal frameworks of many countries, especially in the EU and USA, have been updated to allow the existence of such aggregator mechanisms~\cite{Valarezo2021AnalysisON,Broka2020HandlingOT}.

Nonetheless, there are still many open research topics regarding the functionality and mechanisms of the aggregator. For example, there are privacy concerns about the metering information constantly transmitted between the DERs and the aggregator~\cite{Wagh2020ADP,Gai2020AnED}. Also, there is an interest in designing efficient Demand-Response (DR) aggregator mechanisms to improve the Smart Grid's technical, economic, and market aspects~\cite{Ibrahim2022ARO,Rawat2019ATS}.


In this paper, we employ mechanism design and cooperative game theory ideas to propose a novel aggregator framework for the efficient integration of DERs in the Grid. Our framework provides an aggregation architecture along with mechanisms for its effective and efficient operation and aims to (and, as our experiments show, succeeds in) increase the flexibility offered by the aggregator to the Grid and the profits of the participating agents. 

In our multiagent architecture~\footnote{This is the full version of the paper ``A Novel Aggregation Framework for the Efficient Integration of Distributed Energy Resources in the Smart Grid'' originally presented at AAMAS 2023~\cite{10.5555/3545946.3598986}.}, we introduce the so-called {\em Local Flexibility Estimators (LFEs)} that allow us to address some severe aggregator issues, such as privacy concerns and evaluation of the DERs' flexibility accuracy.
LFEs essentially serve as DER coalition managers, coordinating their members' market activities.   
Given this, our work's focus is the creation of efficient LFE cooperatives intending to increase the profits of every stakeholder. To achieve this, we have populated our framework with various selection mechanisms---some of which are {\em scoring rules}~\cite{Gneiting2007StrictlyPS}, and some are (deep) {\em reinforcement learning (RL)}~\cite{Sutton2005ReinforcementLA} techniques. To the best of our knowledge, using RL for this purpose is entirely novel. An Aggregator agent can then use these selection mechanisms to decide which LFEs to include in its (flexibility) offers to the day-ahead markets. We provide a systematic experimental evaluation using data from the PowerTAC~\cite{Ketter2013PowerTA} simulator in various experimental scenarios that we formulated to test the different aspects of our aggregator framework. Last but not least, our work extends the PowerTAC simulator with aggregator-enabling functionality. 

Arguably, our aggregator framework contributes to the smooth DERs' integration into the Grid since {\em (a)} it allows smaller DERs to participate in the Smart Grid markets; {\em (b)} it selects which LFEs to participate in the energy transactions, increasing the expected accuracy of the promised offers, thus indirectly aiding the Grid's stability; and {\em (c)} as verified via our experiments, the use of certain designed selection and pricing mechanisms leads to higher payments for (at least the competent in terms of prediction accuracy, but also sometimes the not-so-accurate) LFEs that the aggregator manages. Specific findings of our systematic experimentation are:

{\em (1)}~The use of the truthfulness-incentivizing {\em Continuous Ranked Probability Score (CRPS)}~\cite{Gneiting2007StrictlyPS,Robu2012CooperativeVP} mechanism rewards effectively LFEs that have reliable flexibility estimates and results to the highest aggregator-to-LFEs payments for those LFEs, compared to those achieved with other selection mechanisms; or compared to assets' earnings in ``baseline settings'' when they either participate in a ``traditional'' aggregator that manages all available DERs or when they trade directly with the Grid.

{\em (2)}~A {\em Simple Selection} mechanism we put forward ranks as a close second to CRPS. However, this mechanism is easier for non-specialists  to understand. This result implies a trade-off between using a highly efficient yet complex scoring rule vs. a slightly less efficient yet easy-to-understand selection mechanism, since using the latter can motivate the participation of small DERs (e.g., corresponding to small \& medium-sized enterprises or private homes).

{\em (3)}~The {\em DQN Selection} mechanisms were better than the aforementioned baseline settings only for certain settings in which DER accuracy does not fluctuate dynamically over time.

{\em (4)}~Low-accuracy LFEs prefer to participate in larger LFE cooperatives so their errors can be balanced out by the team.
  
{ \em (5)}~When using the {\em CRPS Selection} mechanism, and regardless of the LFEs’ prediction accuracy, our framework results in increased profits for every LFE, compared to those potentially accrued via participation in functional commercial flexibility aggregators paid via pricing mechanism in use in the current Smart Grid~\cite{flexiblepowerPayment}.

\section{Background and Related Work}
This section provides background and reviews related work.


\subsection{Distributed Energy Resources (DERs)}
To begin, DERs, are electricity supply or demand resources located across the Smart Grid. One of the most common DER types are Battery Energy Storage Systems (BESS). These can be either in the form of dedicated batteries or embedded batteries within electric vehicles (EVs)~\cite{Rigas2015ManagingEV,aggregatedFlexibilityEvs}. BESS are essential because they can store energy to shift the demand at other hours of need~\cite{Koufakis2016TowardsAO}; while the eventual incorporation of millions of EVs in the Grid 
creates opportunities for their utilization in demand-response and flexibility aggregation. 
Furthermore, there are many interruptible load users, usually households that offer to limit their electricity usage at specified times to contribute to demand-shifting tasks~\cite{Sahebi2012SimultanousED}. Finally, many types of small-scale renewable energy generators exist, the most common being wind turbines and solar panels that can be located on the rooftops of both apartments and electric vehicles~\cite{Vithayasrichareon2015ImpactOE}.

\subsection{The Role of the Aggregator}

DERs are in general too small to be ``visible'' to the Grid and efficiently participate in energy trading on their own~\cite{Kubli2021BusinessSF,Chalkiadakis2011CooperativesOD}.
Thus the main functionality of an aggregator is to accumulate DER flexibility and perform demand-response and load-balancing to increase the total profit while supporting the Grid stability. This is possible since an aggregator is considered a reliable participant in the energy markets because of the considerable flexibility loads it can control.
 
The current regulations and the state of the energy markets allow existing actors, such as energy suppliers~\cite{Dang2019DistributedGP} and balance responsible parties~\cite{Barbero2020CriticalEO}, to take the role of the aggregator and trade flexibility. This is possible because, in practice, the aggregator's portfolio consists of DER assets usually owned by other stakeholders and utilized by the aggregator to implement its business model. However, there are also independent aggregators~\cite{Schittekatte2021TheRF} which are not affiliated with any other entity like suppliers or balancing utilities. Our framework can support all three types of existing aggregators.

\subsection{PowerTAC: The Smart Grid simulator}
 PowerTAC~\cite{Ketter2013PowerTA} is a rich competitive economic simulation of future energy markets featuring several Smart Grid components(e.g., DERs, retail and wholesale energy markets, etc.). With the help of this simulator, researchers can better understand the behavior of future customer models and experiment with retail and wholesale market decision-making by creating competitive agents and benchmarking their strategies against each other. In this way, a host of useful information is extracted, which can be used by policymakers and industries to prepare for the upcoming market changes.

In our case, we have modified the PowerTAC platform so it can support the addition of aggregator mechanisms, but without altering the behavior of its realistic DER and market models. Another reason we selected this simulation platform is that it employs a realistic day-ahead wholesale market simulating the supply and the demand of future energy markets in a pretty accurate way.

\subsection{Related Work}

Aggregators and their mechanisms have become a very active research topic because of the rise of flexible loads in the grid. Several studies about the business models of aggregators and the way they should operate exist. For example~\cite{Okur2021AggregatorsBM}, reviews the existing aggregator models' operational and economic aspects. Another work~\cite{Lampropoulos2019AFF} proposes a hierarchical control framework that enables the provision of flexible services in power systems through aggregation entities. Zheng et al. ~\cite{Zheng2021AnAR} have developed an aggregator-based resource allocation system using an artificial neural network and other optimization techniques. Finally,~\cite{tesi_carducci} proposes, among other things, a flexibility aggregation architecture that uses so-called ``minimum flexibility units'' that operate at a local level and represent single flexibility assets, industrial microgrids, or multiple end users. These units however always correspond to a single meter and do not manage heterogeneous DERs nor evaluate their accuracy.

Furthermore, there are many  works about the demand response operations of the aggregators~\cite{Ibrahim2022ARO,Gkatzikis2013TheRO,goflex}. SEMIAH~\cite{Jacobsen2015SEMIAHAA} is an aggregator framework designed to support European demand response programs. It uses a component-based architecture and focuses on the functionality of the virtual power plant. Additionally, there are distributed algorithms developed for large-scale demand response aggregation~\cite{CasteloBecerra2017CooperativeDA,Mhanna}. Also, Rawat et al. ~\cite{Rawat2019ATS} propose a two-stage interactive, responsive load scheduling model developed between a demand response aggregator and the distribution system operator.

One more critical aspect of the aggregator research is the security of information transactions and the integration of electric vehicles. Wagh et al.~\cite{Wagh2020ADP} address the problem of cybersecurity by developing a distributed privacy reserving framework that aggregators can adapt. Another study~\cite{Gai2020AnED} with a similar research focus proposes an aggregation scheme with local differential privacy that can efficiently and practically estimate power supply and demand statistics while preserving any individual participant's privacy. Finally, there are many studies about the aggregation of EVs~\cite{Shafiekhah2016OptimalBO}. Specifically, many of these studies focus on the charging scheduling of EVs~\cite{Goyal2016CustomerAA,Vagropoulos2016RealTimeCM,GonzlezVay2016SelfSO}. 
Other works study the payment mechanisms for electric vehicle aggregators~\cite{PerezDiaz2018CoordinationAP,PerezDiaz2018CoordinationOE}. 
Last but not least, several works in the 
literature approach the aggregation problem as a coalition formation one, proposing game-theoretic and mechanism design solutions to form DER cooperatives so that these are able to participate in demand-response tasks~\cite{Akasiadis2016DecentralizedLE,Chalkiadakis2011CooperativesOD, Robu2012CooperativeVP}.
Interestingly, \cite{Akasiadis2016DecentralizedLE} demonstrates that creating larger coalitions, such as the LFEs in our case, can lead to more efficient energy trading.


\section{A Novel DER Aggregator Framework}

\begin{figure}[!ht]
  \centering
  \includegraphics[width=1\linewidth]{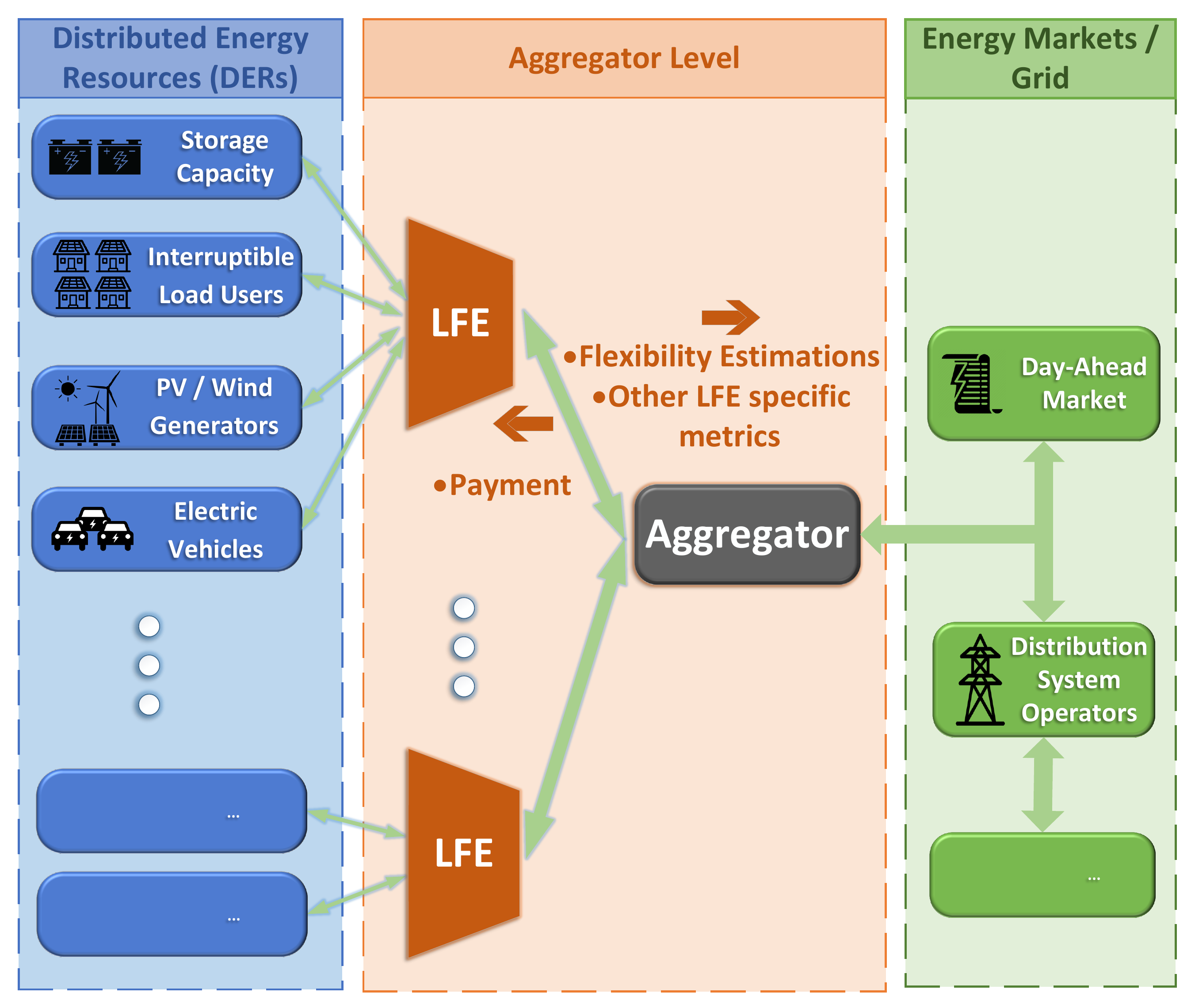}
  \caption{ The component diagram of the proposed aggregator architecture consists of three levels; the Distributed Energy Resources, the Aggregator Level, and the Energy Markets }
  \label{fig:framework}
  \Description{The component diagram of the proposed aggregator architecture consists of three levels; the Distributed Energy Resources, the Aggregator Level, and the Energy Markets}
\end{figure}

Privacy concerns, impartial and fair scoring of the LFEs, and efficient management of all stakeholders' assets are the main problems this two-step DER aggregation architecture addresses. In particular, our framework provides an {\em aggregation architecture} along with {\em mechanisms} for its effective and efficient operation, increasing the actual flexibility offered by the aggregator to the Grid and the profits of the participating agents. Additionally, the proposed architecture can be distinguished into three levels, as depicted in Fig.~\ref{fig:framework}. The first is the \emph{Distributed Energy Resources} level consisting of many Smart Grid assets. Specifically, our framework supports smaller and bigger scale DERs utilized by the Smart Grid. The second level of this framework is named the \emph{Aggregator Level}, and it consists of Local Flexibility Estimators (LFEs) organizing the DERs into coalitions, and a central Aggregator agent which selects which LFEs to include in a cooperative to participate in flexibility trading in the Smart Grid. The third level of this framework represents the Smart Grid itself---specifically, the energy markets to which the Aggregator-coordinated LFEs cooperative participates.
We now proceed to describe the key parts of our overall framework in detail.

\subsection{Local Flexibility Estimators (LFEs)}
An LFE agent
acts as a coordinator of a coalition consisting of a varying number of heterogeneous DERs, effectively offering them visibility to the Grid. This means that all DER assets, regardless of size, can now participate indirectly in a 
flexibility aggregation process: though originally potentially too 
small to bid in the energy markets directly~\cite{Kubli2021BusinessSF}, 
DERs can now form LFE-coordinated coalitions; 
which in turn can be selected by the Aggregator to participate in LFE cooperatives to trade the accumulated flexibility. 

The rationale and particular method an LFE uses to select its participating DERs are not absolute, as it depends on a multitude of constraints, priorities, or other factors.  One factor could be locality limitations. For instance, it might be easier to pick all DER assets lying close to each other in a physical neighborhood because of how the smart meters are placed~\cite{Kabalci2016ASO}. Alternatively, an LFE might be formed for the convenience of or to serve the requirements of a single stakeholder or a group of stakeholders with the same goals, like Local Energy Communities~\cite{Lezama2019LocalEM} or small companies. 
LFEs can then employ any cooperative formation method of their choosing that respects the stakeholders' requirements; for instance, they can use the very same member selection mechanisms that an Aggregator may employ, which we describe in detail in Section~\ref{sec:selectionmechanisms}.

One of the primary responsibilities of an LFE is to monitor the consumption and production of the DERs it controls. Importantly, it can then use the historical consumption/production data to generate accurate estimations of the total flexibility for all participating DERs. In addition to the flexibility predictions, an LFE also provides the aggregator with other necessary metrics, such as confidence in the predictions (see Section~\ref{sec:selectionmechanismsCRPS}).
As a result, each LFE handles a portion of the total aggregator flexibility estimation problem; LFEs provide the aggregator with all information necessary for it to participate in the market, thus reducing the computational complexity of the aggregator optimization process~\cite{Wang2017WirelessBD}.
Note that expanding the functionality of existing DER agents to ease the aggregator's computational burden is not an option in most practical settings, since DERs (EVs, PV, wind turbines) usually if not always come with given properties and capabilities; nevertheless, all DERs have communication capabilities, and thus can participate in LFEs which take up the additional optimization tasks.

Most importantly, LFEs are key to satisfying privacy requirements: all the DER-related information is constrained and protected under the LFEs, so only the corresponding LFE can access the private data of each DER. Additionally, the proposed distributed aggregator scheme enables the usage of Smart Grid blockchain technologies~\cite{Mnatsakanyan2020BlockchainIntegratedVP,Kursawe2011PrivacyFriendlyAF}, which have been used for Smart Grid applications to secure the transactions made, while also protecting the private information required for the transactions. 

Summing up, LFEs are end-nodes that only have to publish to the Aggregator limited information with respect to their flexibility and availability. Other works try to preserve the privacy of DERs by introducing new communication protocols and algorithms~\cite{Wagh2020ADP,Gai2020AnED}. We propose a new all-in-one framework that can, on top of all other benefits, contribute to the privacy concerns that arise when information flows from DERs toward the Aggregator. For instance, there are commercial aggregators that group DERs daily in a relatively simple, straightforward manner, but DERs have to share operation patterns (i.e., private information) with the Aggregator. In our case, DERs with common interests could form an LFE and participate as a multipurpose flexibility entity in any Aggregator, hiding their details from the (potentially non-trusted) Aggregator.

\subsection{The Aggregator agent}

The aggregator in our framework possesses all the properties of a usual Smart Grid aggregator~\cite{Gkatzikis2013TheRO}; however, it has some additional novel properties too, such as the ability to form LFE cooperatives. The framework we propose is comprised of a single aggregator that directly controls the assigned LFEs and indirectly manages all the DERs that the LFEs closely monitor. Additionally, the aggregator is responsible for gathering all the smart-metering data, e.g., flexibility estimations already pre-processed by the LFEs, to efficiently manage the assets and trade their excess flexibility. To address the problem of increasing the profits of each stakeholder, our Aggregator deploys a variety of scoring rules so it can fairly rank the LFEs according to their historic flexibility estimations' accuracy. 

An important benefit of incorporating LFEs is that it enables the Aggregator agent to use formation mechanisms, so as to create efficient LFE cooperatives to increase profits, while also incentivizing truthful and accurate predictions contributing to Grid stability.
Ranking and selecting LFEs is key, since LFEs with unreliable (low-accuracy predictions regarding their) flexibility estimations should not participate in the Aggregator's flexibility offers since they can damage both the Aggregator's profits and overall reputation. 
This does not mean that low-accuracy LFEs will be excluded from the markets. Instead, they can trade with the Grid directly.\footnote{Notice that though LFEs can trade with the Grid directly, the distinct Aggregator architectural level is required, because otherwise, it would not be possible for non-local DERs to group together and reap the benefits of providing large, and potentially highly diverse, flexible loads.}

The Aggregator is thus able to calculate the total energy flexibility it can offer by selecting which LFEs will participate in the upcoming flexibility trades, using scoring and ranking mechanisms such as the ones we propose below.
Moreover, the Aggregator is responsible for splitting the profits back to the LFEs, based on their contribution and appropriate scoring mechanisms that may also take into account the accuracy of LFE flexibility predictions.

The total communication complexity of the proposed framework certainly increases compared to a traditional aggregator~\cite{Gkatzikis2013TheRO}. However, notice that LFEs are local (e.g., managing the assets of a single company or local energy community); hence the added communication load with their DERs is expected to be minimal.

Overall, incorporating LFEs within an Aggregator agent allows us to generalize and scale up, since an Aggregator can serve more DERs by just adding LFEs to take up some of its optimization processes. Therefore, offloading some of the optimization complexity to a lower level (that of LFEs) could mean more accurate and scalable outcomes in terms of flexibility provided to the Grid.

\subsection{Flexibility Estimation}

In a Smart Grid context, total aggregator flexibility is its capability of shifting electrical loads either from itself to the Grid, or in the opposite direction~\cite{Lampropoulos2019AFF}; and there are many works that deal with flexibility estimation or forecasting~\cite{Iraklis2021FlexibilityFA,flexforecast}. At a given timeslot $t$, we calculate the flexibility provided by each DER as follows.

To begin, BESS's available flexibility is proportional to its current energy level and specific charge/ discharge speed in KWh (Eq.~\ref{eq:flex_batteries}).
\begin{equation}
    flex_{BESS}(t) = Charge (or\:Discharge)\;Speed(t)
    \label{eq:flex_batteries}
\end{equation}

 The same principle applies to EVs,  
with the difference that to calculate EVs' flexibility $flex_{EV}(t)$, one has to account for a minimum battery level they should  maintain in order to continue traveling.


The flexibility that interruptible load users can provide was set (following ~\cite{Iraklis2021FlexibilityFA}), to 10\% of the load they are currently using. Hence, upon request of the aggregator through the LFEs, interruptible load users can alter their energy consumption by 10\%:

\begin{equation}
    flex_{InterruptibleLoadUsers}(t) = 10\%\:Load(t)
    \label{eq:flex_ir}
\end{equation}

Most {\em renewable energy-producing} DERs do not have the ability to halt their production, so the flexibility of such DER assets is represented by the amount of energy they produce (Eq.~\ref{eq:flex_prod}):

\begin{equation}
    flex_{EnergyGenerators}(t) = Energy\;Generated(t)
    \label{eq:flex_prod}
\end{equation}

Then, the total flexibility of an LFE is the aggregate flexibility of all DER assets it controls. Hence, the  flexibility of $LFE_i$ that controls a set $K$ of DERs at timeslot $t$ is defined as follows (Eq.~\ref{eq:flex_total}):

\begin{equation}      
    flex_{LFEi}(t) = \sum_{\forall k \in K} flex_{DER_k}(t)
    \label{eq:flex_total}    
\end{equation}
where $flex_{DER_k}(t)$ is calculated given the DER's type.

Finally, the total flexibility of the Aggregator is calculated as in Eq.~\ref{eq:flex_agg} below, where $S$ is the set of LFEs selected by the Aggregator to contribute at timeslot $t$.
\begin{equation}      
    flex_{Agg}(t) = \sum_{\forall s \in S} flex_{LFE_s}(t)
    \label{eq:flex_agg}    
\end{equation}

\subsection{Selection mechanisms}
\label{sec:selectionmechanisms}

A key problem our framework addresses is the formation of efficient LFEs cooperatives to participate in the 
flexibility trading. 
To facilitate this in an impartial manner, the Aggregator first ranks the LFEs using certain scoring functions. 
All our proposed scoring methods calculate the $Score_{LFEi}$ for each $LFE_i$ of the Aggregator, regardless of their prior participation in the latest flexibility tradings of the Aggregator. The Aggregator needs to score the LFEs regularly to have accurate information regarding the performance of every LFE it controls. Additionally, it is possible that during some specific periods of the year, the accuracy of the flexibility predictions can fluctuate~\cite{Bosman2016DifficultiesAR}, so this can also be an essential factor in the aggregator's decision process. 

In our experiments, the Aggregator selects the LFEs with the $LFE_i$ scores that exceed an Aggregator-specified threshold, in order to calculate its $flex_{Agg}(t)$ flexibility at time $t$.
We now present our scoring mechanisms in detail.

\subsubsection{Simple Selection mechanism}
\label{simple-selection-mech}
The first selection mechanism we developed uses the Mean Absolute Error (MAE) of the $\widetilde{flex}_{LFEi}$ flexibility prediction. Specifically, for an $LFE_i$ that estimates its flexibility for the next $k$ hours, we calculate the MAE as in Eq.~\ref{eq:mae}, Similarly, we define the average flexibility of an LFE as in Eq.~\ref{eq:avg_flex}.
\begin{equation}      
    MAE_{LFEi}(t) = \frac{\sum_{j=1}^{k} |\widetilde{flex}_{LFEi}(j) - flex_{LFEi}(j)|}{k}
    \label{eq:mae}    
\end{equation}
\begin{equation}      
    AvgFlex_{LFEi}(t) = \frac{\sum_{j=1}^{k} |flex_{LFEi}(j)|}{k}
    \label{eq:avg_flex}    
\end{equation}

Then, we calculate the $Score_{LFEi}(t)$ for each $LFE_i$ of the Aggregator by using  Eq.~\ref{eq:simple_scoring}. The first step is to divide the $MAE_{LFEi}(t)$ by the average actual flexibility, then we subtract that value from $1$ to have a straightforward confidence percentage. Finally, to avoid negative scores, we bound the score value, ${Score_{LFEi}(t)\in[0,1]}$.

\begin{equation}      
    Score_{LFEi}(t) = max\left( 1 - \frac{MAE_{LFEi}(t)}{AvgFlex_{LFEi}(t)},0\right)
    \label{eq:simple_scoring}    
\end{equation}

\noindent Thus, the score is at maximum when the predictor is perfect ($LFE_i$ has an MAE of $0$). By contrast, the score is $0$ when the predictions' mean absolute error is worse than that of having a ``dummy'' predictor that always outputs $0$ values, i.e., when the prediction MAE is greater than the average flexibility. In the end, the aggregator calculates the average score of $LFE_i$ over a time window $w$ of past trading cycles (Eq.~\ref{eq:avg_simplescore}). Then the aggregator checks if the average score $AvgScore_{LFEi}(t)$ of $LFE_i$ is over the desired threshold $\tau$. The central concept of this metric is to get a simple and easy-to-compute confidence percentage so that we can use it as a ``naive'' selection method to compare with other, more sophisticated solutions.

\begin{equation}
    AvgScore_{LFEi}(t) = \frac{\sum_{l=t-w}^{t} MAE_{LFEi}(l)}{w}
     \label{eq:avg_simplescore}  
\end{equation}

\subsubsection{Continuous Ranked Probability Score (CRPS)}
\label{sec:selectionmechanismsCRPS}
The second selection mechanism we deployed uses the Continuous Ranked Probability Score (CRPS)~\cite{Gneiting2007StrictlyPS}, which assesses the accuracy of a probabilistic prediction over the actual occurrence. CRPS has been previously used  for virtual power plant formation~\cite{Robu2012CooperativeVP}, and we use it in our aggregator framework to score the LFEs predictions aiming to optimize aggregator profits. CRPS is a strictly proper scoring rule, meaning that the expected score is maximized only if predictors accurately report their expectation over the prediction error they can potentially make~\cite{Gneiting2007StrictlyPS}. CRPS has been shown to incentivize energy suppliers to be truthful and accurate~\cite{Robu2012CooperativeVP}, and we use it here to incentivize truthful and accurate LFE flexibility predictions.

With the {\em Simple Selection} mechanism, the LFEs only had to send their flexibility predictions to the aggregator. When using the {\em CRPS Selection} mechanism, LFEs also need to provide the uncertainty over the prediction error, and are rewarded accordingly: estimates that are both accurate and highly confident will be the ones achieving higher CRPS scores. 
The $CRPS_{LFEi}(t)$ is defined as follows: 
\begin{equation}      
    CRPS_{i}(t) = \sigma(t) \left[\frac{1}{\sqrt{\pi}} - 2 \phi\left( \frac{e_i(t)}{\sigma(t)}\right)- \frac{e_i(t)}{\sigma(t)}(2\Phi\left(\frac{e_i(t)}{\sigma(t)}\right) -1)\right]
    \label{eq:crps}    
\end{equation}
where $\phi$ and $\Phi$ denote the probability density and the cumulative distribution function of a {\em standard Normal} variable.
In our case, $e_i$ is the relative prediction error, as shown in Eq.~\ref{eq:error}. 
\begin{equation}
    e_i(t) = \frac{{flex_{LFEi}(t)}- \widetilde{flex}_{LFEi}(t)}{ \widetilde{flex}_{LFEi}(t)}
    \label{eq:error}    
\end{equation}
(which we ensure to take values in [-1,1]).
Additionally, along with every prediction, the LFEs also send a Normal distribution $\mathcal{N}(0,\sigma(t)^2)$ to describe their uncertainty over their error, where 
$\sigma(t)$ changes over time according to the LFEs uncertainty on the predictions. (We assume, as in~\cite{Robu2012CooperativeVP}, that random
 errors, over a long enough period, will be normally
distributed around a mean of 0.)

\begin{equation}
    AvgScore_{LFEi}(t) = \frac{\sum_{l=t-w}^{t} CRPS_i(l)}{w}
     \label{eq:avg_crpsscore}  
\end{equation}

Similarly to the {\em Simple Selection} mechanism, the aggregator calculates the average score of each $LFE_i$ over a time window $w$ resembling past trading cycles (Eq.~\ref{eq:avg_crpsscore}). Then the aggregator checks if the average score of $LFE_i$ is over a threshold $\tau$.

\subsubsection{Reinforcement Learning: DQN}
\label{sectionRL}
The final selection mechanism we developed deploys the celebrated Q-Networks (DQN)~\cite{Mnih2013PlayingAW} RL algorithm. We formulate the aggregator's decision-making problem as a decision process, aiming to find the action with the highest Q-value---corresponding to the long-term utility of taking a (selection) action, i.e., selecting a set of LFEs at this time step. We assume a continuous state-space defined via providing as inputs the $CRPS_{LFEi}(t)$ and the $\widetilde{flex}_{LFEi}(t)$ for every LFE of the aggregator; and define a discrete action-space containing all the different possible selections of LFEs, using one-hot encoding.
For example, for $n$ LFEs, there will be $n$ bits of information, with $1$ meaning the LFE was selected, and $0$ the opposite, so when the aggregator selects among $n$ LFEs, the action-space will have $2^n$ possible actions. 
\begin{figure}[ht]
  \centering
  \includegraphics[width=1\linewidth]{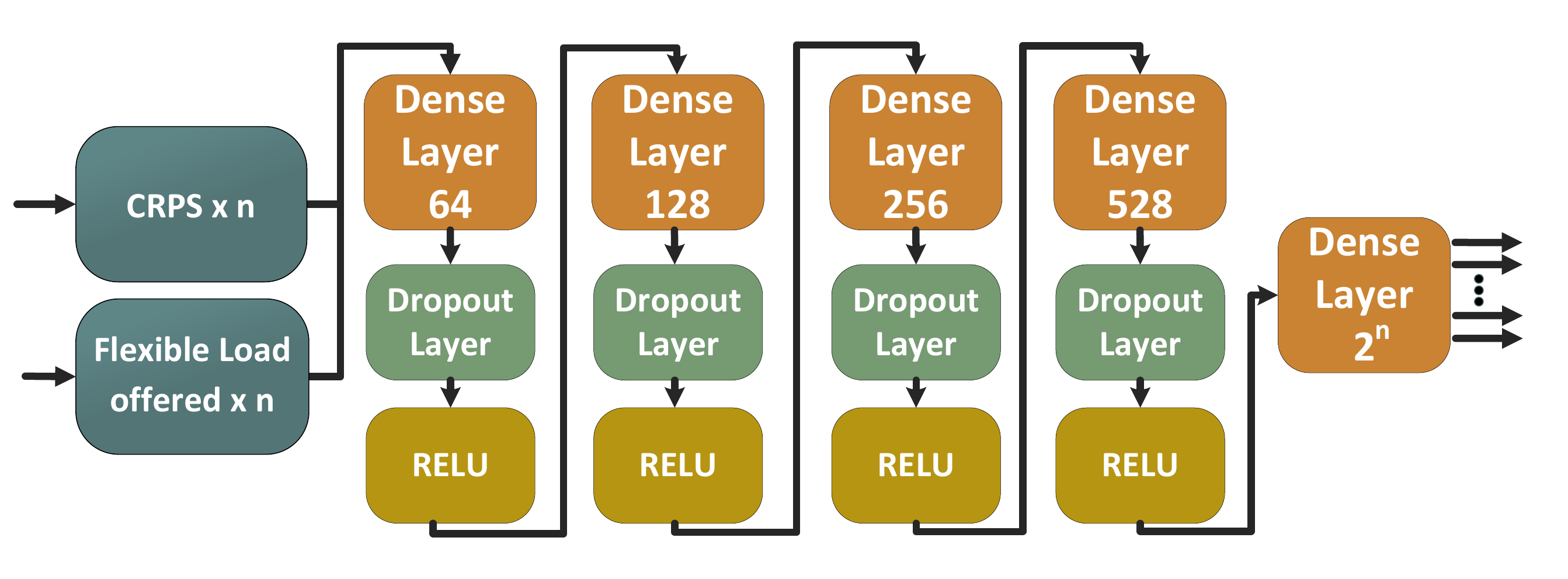}
  \caption{The architecture of the Reinforcement Learning model used for the selection of the best LFE cooperative.}
  \label{fig:rl_model}
  \Description{The architecture of the Reinforcement Learning model used for the selection of the best LFE cooperative}
\end{figure}

DQN is a value-based method that does not store any explicit policy but only a value function in the form of a deep neural network, hence the policy is implicit and can be derived directly from the neural network as the action with the highest value. In Fig.\ref{fig:rl_model}, we can see our neural network architecture. We used five fully-connected layers with increasing nodes, so we could better handle this complex optimization problem. Also, we put three "Drop-Out" layers followed by rectified linear activation units (ReLU) to address the randomness of the samples during training.

Two reward functions were developed for the training of the DQN model. The goal of the first reward function is to maximize the profits of the aggregator during the evaluation time (Eq.~\ref{eq:reward1}). $Z$ is a normalization constant that keeps the reward values closer to $0$, but without an upper bound. The term $V_{Grid,Agg}(t)$ refers to the Grid payments towards the Aggregator for the flexibility trading it has completed in timeslot $t$. 

\begin{equation}      
    Reward_{1}(t) = \frac{V_{Grid,Agg}(t)}{Z}
    \label{eq:reward1}    
\end{equation}

The second reward function we tested tries to balance the profits of both the Aggregator and the LFEs (Eq.~\ref{eq:reward2}). This function takes values in the $[0,1]$ interval. Here too, $J$ is the set of LFEs selected to participate in the aggregator's latest flexibility trades, and $S$ is a set comprised of every LFE in the aggregator. We define  $V_{Agg,LFEi}(t)$ to refer to Aggregator payments to each $LFE_i$ for participating in its flexibility trading. Also, we define $V_{Grid,LFEi}(t)$ to refer to the payments received by $LFE_i$ at $t$ while trading flexibility directly with the Grid (when not selected by the Aggregator in timeslot $t$). 

\begin{equation}      
    Reward_{2}(t) = \frac{V_{Grid,Agg}(t)}{ V_{Grid,Agg}(t) + \sum_{\forall i \in \{S-J\}} V_{Grid,LFE_i}(t) }
    \label{eq:reward2}    
\end{equation}

\subsection{Pricing Mechanisms}
\label{sec:pricingmechanisms}
Another problem our framework addresses is the 
distribution of the aggregators' profits to each individual LFE. We have deployed two different pricing mechanisms inspired by previous studies. At first, using these mechanisms, we calculate the payment from the energy markets to the aggregator for trading its flexibility and then 
payments by the aggregator to its 
LFEs.
The (per KWh) price of the traded energy  in the markets at timeslot $t$ is denoted as $p(t)$.
As mentioned earlier, PowerTAC simulates the demand and supply of a day-ahead wholesale market, which means that the price of the KWh, denoted as $p(t)$ below, changes dynamically through the course of the simulations.

\subsubsection{Prediction Accuracy mechanism}
\label{predaccuracy}
Our first pricing mechanism calculates the payments based only on ``point estimates'' of the accuracy of the flexibility predictions (i.e., no uncertainty-related distribution is reported), as in~\cite{Chalkiadakis2011CooperativesOD}. The more accurate the flexibility estimators, the higher the payments it awards them.

To begin, Eq.~\ref{eq:grid_agg_simple} shows how the Aggregator is rewarded for trading its flexibility in the energy markets. 

\begin{equation}      
    V_{Grid,Agg}(t) = \frac{log|flex_{Agg}(t)| \cdot flex_{Agg}(t) }{1+\alpha\cdot e
    _{Agg}(t)^\beta} \cdot p(t)
    \label{eq:grid_agg_simple}    
\end{equation}
The logarithmic term increases with the provided flexibility. This incentivizes the Aggregator to include a large number of LFEs in its offers. At the same time, the Aggregator has to proceed in its LFEs selection with caution, as its flexibility prediction error, denoted by $e_{Agg}$ and calculated as in Eq.~\ref{eq:error}, plays a role in its final reward:
notice that the parameters of the denominator resemble a bell-shaped function, so the value is maximized when the prediction error is zero.  
Parameters $\alpha$ and $\beta$ determine the exact shape of the curve~\cite{Chalkiadakis2011CooperativesOD}; in our experiments, we set $\alpha = 1.6$ and $\beta = 4$.
The LFEs not selected by the Aggregator trade directly with the Grid, and are also rewarded according to Eq.~\ref{eq:grid_agg_simple} by swapping the terms $e_{Agg}(t)$ with $e_{LFEi}(t)$ and $flex_{Agg}$(t) with $flex_{LFEi}$(t).

After the Aggregator is paid, it distributes the profits to the contributing LFEs. Eq.~\ref{eq:agg_lfe_simple} displays the pricing function used to distribute $V_{Grid,Agg}$ to the LFEs based on their prediction error $e_i$.

\begin{equation}      
    V_{Agg,LFE_i}(t) = \frac{Z}{1+\alpha\cdot e_{i}(t)^\beta} \cdot \frac{flex_{i}(t)}{flex_{Agg}(t)} \cdot  V_{Grid,Agg}(t) 
    \label{eq:agg_lfe_simple}    
\end{equation}
The term $flex_{i}(t)/flex_{Agg}(t)$ represents the flexibility contribution percentage of the $LFE_i$ to the Aggregator. Also, $Z$ is a normalization parameter calculated to split the payment to every LFE completely. 
One can adjust $Z$ to allow for some portion of the total payment to be withheld by the Aggregator. 

\subsubsection{CRPS based mechanism}
\label{sec:crps_payments}
The second pricing mechanism~\cite{Robu2012CooperativeVP} uses the CRPS score. This mechanism encourages the formation of larger LFE cooperatives while giving incentives for accurate and truthful flexibility predictions. A key difference with the Prediction Accuracy-only pricing mechanism, is that the CRPS-based mechanism can be more forgiving to low-accuracy predictors. This is because LFEs also provide their flexibility predictions' uncertainty distributions, 
accounted for in their CRPS scores (see Sec.~\ref{sec:selectionmechanismsCRPS}).

The "Grid-to-Aggregator" payment shown in Eq.~\ref{eq:grid_agg_crps} is calculated similarly to the previous mechanism. However, the accuracy factor is the aggregator's CRPS value (normalized to $[0,1]$ similarly to what is done in~\cite{Robu2012CooperativeVP,ROBU201619}) instead of a bell-shaped function. 
\begin{equation}      
    V_{Grid,Agg}(t) = CRPS_{Agg}(t) \cdot log|flex_{Agg}(t)| \cdot flex_{Agg}(t) \cdot p(t)
    \label{eq:grid_agg_crps}    
\end{equation}

Here too, the LFEs that were not selected by the aggregator use Eq.~\ref{eq:grid_agg_crps} but with the relevant $LFEi$ terms to calculate their "Grid-to-LFE" payment. After the aggregator is paid, it distributes the profits to each selected LFE member (set $J$ of LFEs) using Eq.~\ref{eq:agg_lfe_crps}. 

\begin{equation}      
    V_{Agg,LFE_i}(t) = \frac{CRPS_{Agg}(t) \cdot flex_{LFEi}(t) \cdot V_{Grid,Agg}(t)}{\sum_{\forall j \in J} \left(CRPS_{LFEj}(t) \cdot flex_{LFEj}(t)\right)}
    \label{eq:agg_lfe_crps}    
\end{equation}

This pricing mechanism ensures that each participant is awarded a weighted portion of the total payment based on their contribution and their individual CRPS score. Additionally, the CRPS is helpful because it shows how beneficial the LFE estimates were for the total flexibility trading of the aggregator.


\subsubsection{Simple Grid-to-Aggregator payment}
The use of CRPS in pricing mechanisms (Sec.~\ref{sec:crps_payments}) has been proposed in Smart Grid research  because CRPS promotes predictors' truthfulness and accuracy, punishing as it does untruthful and inaccurate predictors, irrespective of their size~\cite{Robu2012CooperativeVP,ROBU201619}. However, currently the CRPS mechanism is not utilized for the calculation of payments in real-world energy trading. Instead, modern energy markets use simple mechanisms that are exclusively based on the final flexibility contribution, and not on the accuracy of the prediction~\cite{flexiblepowerPayment}. The standard Grid payments are calculated (Eq.~\ref{eq:simple_payment}) using the promised (predicted) flexibility $\widetilde{flex}_{i}$, the actual volume of delivered flexibility, $flex_{i}$, the pre-agreed electricity price $p(t)$, and the current (at delivery) electricity price $p_{c}(t)$. The difference between the predicted and the actual flexibility is defined as $flex_{\diff}(t) = flex_i(t) - \widetilde{flex}_i(t)$. 

\begin{equation}
V_{Grid,i}(t) = \begin{cases}
			 flex_i(t) \cdot p(t) & flex_{\diff}(t)\leq 0\\
             \widetilde{flex}_i(t) \cdot p(t) + flex_{\diff}(t) \cdot p_{c}(t) & flex_{\diff}(t) > 0
		 \end{cases}
   \label{eq:simple_payment}    
\end{equation}

With this payment type, the Aggregator is rewarded with a pre-agreed electricity price $p(t)$ when it delivers (up to) a promised amount of flexibility at $t$ timeslot; and is paid according to the current energy price for any additional flexibility amount it delivers.

\section{Experimental Evaluation}

Here evaluate the performance of the LFEs and the Aggregator in various scenarios formulated to test different aspects of our framework. We used five different LFE selection methods, with four variants each, combining {\em static} vs. {\em dynamic LFE accuracy}, and {\em CRPS} vs. {\em Simple Grid-to-Aggregator payment}.
\begin{figure}[ht]
  \centering
  \includegraphics[width=1\linewidth]{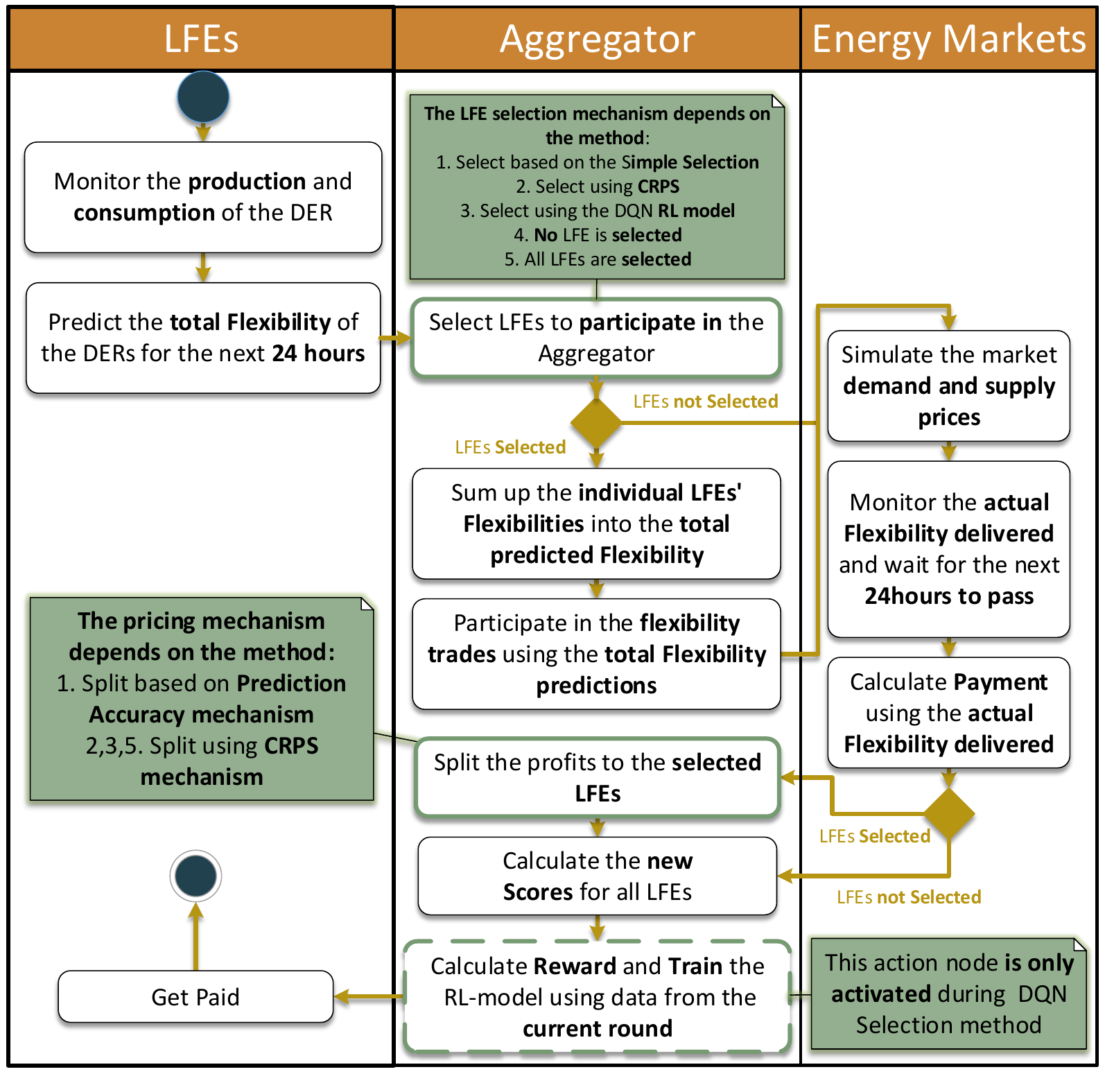}
  \caption{Trading flexibility via various methods.}
  \label{fig:scenarios}
  \Description{Trading flexibility via various methods.}
\end{figure}

To lay the ground for our experiments, we first provide a detailed activity diagram (Fig.~\ref{fig:scenarios}) depicting the actions taken by the LFEs and the Aggregator to trade flexibility with the Grid over a 24-timeslot period starting from 12:00 pm every day. 
As we can see, the main activity flow remains the same for all experimental methods, but some critical variations will be discussed explicitly later.

First, the LFEs monitor and process the production and consumption data of all their DERs, to train their flexibility estimators. At the start of every trading cycle, the LFEs deliver their flexibility predictions for the next 24 hours to the Aggregator, to trade in the day-ahead wholesale market. 
Then the Aggregator chooses which LFEs to participate in the upcoming flexibility trades using a method-specific selection mechanism. Afterward, the Aggregator calculates its total flexibility estimate and participates in the flexibility tradings. At the same time, PowerTAC simulates the demand and supply prices, so that energy market participants can buy or sell energy. Once the day-ahead market closes, PowerTAC continues to monitor the actual flexibility delivered until the market opens again. Then, it calculates the aggregator's payment based on the delivered flexibility and the prediction error, as seen in Section~\ref{sec:pricingmechanisms}. We make the assumption that all energy offered to the Grid is always bought.
After the Aggregator has received its payment, it distributes the profits to the LFEs via a pricing mechanism. Meanwhile, the LFEs not selected by the  Aggregator  have also traded their flexibility directly with the energy markets. The final step for the Aggregator is to assess the prediction accuracy of every LFE, selected or not, by re-calculating the {\em Simple Score} and the {\em CRPS} metrics. After each trading cycle, one has to re-evaluate to account for possibly fluctuating LFE prediction performance, so the aggregator will know which LFEs to select for subsequent trades. 

The total number of DERs simulated in PowerTAC resembles that of a small city; hence, there are considerable amounts of available flexible loads to trade: specifically, a total of 13 MWh of renewable energy can be produced, and a total of 14 MWh can be consumed at peak hours respectively. Also, there is a total storage capacity of 7.5 MWh, where BESS are accountable for 2.5 MWh while the rest derives from EVs. We divided the PowerTAC's DER assets {\em randomly} into twelve {\em heterogeneous} LFEs consisting of various DER assets such as solar panels, BESS, electric vehicles, households, and interruptible consumption users. The LFE heterogeneity makes the simulations much more realistic since every LFE has different attributes. Some of these attributes refer to the electricity consumption/production profiles and the total storage capacity.

We use PowerTAC's day-ahead wholesale market to trade flexibility in our experiments. Specifically, a simulated day is divided into 24 timeslots resembling the hours of the day. Like in many 
real
day-ahead energy markets, in our 
experiments 
all offers should be submitted before 12:00 pm, so there is a need for each LFE to provide the Aggregator with at least 24-hour-ahead flexibility predictions. 
The LFEs' flexibility predictions are the outcome of a Gaussian 
random process per $LFE_i$, $\mathcal{G}(\mu_{i},\sigma_{i})$, with $\mu_{i}$ being the actual $LFE_i$ flexibility ($flex_{LFEi}$), and variance $\sigma_{i}$ an $LFE_i$-specific parameter. 

\subsection{Methods Instantiation}  


As mentioned, we have formulated five methods corresponding to different LFE selection and pricing mechanisms (distributing the Aggregator's profits to the LFEs):

\subsubsection{Using Simple Selection}
The first method uses the \emph{Simple Selection mechanism} to decide which LFEs to participate in the aggregator. In detail, the aggregator calculates the average score of $LFE_i$ over a time period $w=3$ of past trading cycles as displayed in Sec.~\ref{simple-selection-mech}. In our experiments, we set threshold $\tau = 0.7$. Also, we use the \emph{Prediction Accuracy} pricing mechanism for this method as described in Sec.~\ref{predaccuracy} since we assume that LFEs in this setting are unable to provide distributions over their prediction error. We employed the {\em CRPS Pricing} mechanism for all other experimental methods since it provides incentives for truthful and reliable LFE predictions~\cite{Robu2012CooperativeVP, ROBU201619}, which makes it a perfect fit for our framework.

\subsubsection{CRPS Selection}
The second method uses the more sophisticated {\em CRPS Selection}. Here too, we calculate the average CRPS score of a time period $w=3$ and check if the final CRPS score is higher than a specific threshold (see Sec.~\ref{sec:selectionmechanismsCRPS}). Higher (normalized to $[0,1]$) CRPS values represent LFEs with higher prediction accuracy, while lower CRPS values correspond to less accurate LFEs. Furthermore, we set the threshold $\tau = 0.77$ since this was observed empirically to achieve similar selection rates (numbers of LFEs selected per timeslot) with the previous mechanism.

\subsubsection{DQN Selection}
This method evaluates {\em DQN Selection} using two different reward functions (see Sec.~\ref{sectionRL}).
Note that here the aggregator also has to calculate the RL reward and train the RL selection model using  information from the latest trading cycle.

\subsubsection{Singleton LFEs}
In this method, as the name suggests, every LFE interacts directly with the energy markets using the {\em CRPS Pricing} mechanism. This use case acts as the first baseline method depicting how the LFEs would have performed if they had never participated in our aggregator framework.

\subsubsection{All LFEs}
Here all LFEs participate in the Aggregator without any selection criteria. This use case mirrors the current state-of-the-art aggregation scenario, in which an aggregator incorporates all available resources at hand.

\subsection{Experimental Scenarios}
We composed four experimental scenarios to test the performance of our framework in different use cases. As depicted in Table~\ref{tab:config_scenarios}, the scenarios are dependent: (a) on the stability of the LFE's prediction accuracy throughout the simulations, (b) on how the Aggregator agent is compensated for providing flexibility services to the Grid.

\begin{table}[!ht]
\begin{tabular}{|c|cc|cc|}
\hline
           & \multicolumn{2}{c|}{LFE Accuracy}     & \multicolumn{2}{c|}{\begin{tabular}[c]{@{}c@{}}Grid-to-Aggregator\\  Payment\end{tabular}} \\ \hline
\textbf{Scenarios}  & \multicolumn{1}{c|}{Static} & Dynamic & \multicolumn{1}{c|}{CRPS}                             & Simple                             \\ \hline
Scenario 1 & \multicolumn{1}{c|}{\checkmark}      &         & \multicolumn{1}{c|}{\checkmark}                                &                                    \\ \hline
Scenario 2 & \multicolumn{1}{c|}{}       & \checkmark       & \multicolumn{1}{c|}{\checkmark}                                &                                    \\ \hline
Scenario 3 & \multicolumn{1}{c|}{\checkmark}      &         & \multicolumn{1}{c|}{}                                 & \checkmark                                  \\ \hline
Scenario 4 & \multicolumn{1}{c|}{}       & \checkmark       & \multicolumn{1}{c|}{}                                 & \checkmark                                  \\ \hline
\end{tabular}
\caption{Configuration of the four experimental scenarios.}
\Description{Configuration of the four experimental scenarios.}
\label{tab:config_scenarios}
\end{table}

We have formed 12 LFEs consisting of various heterogeneous DERs for our experiments. When the LFE prediction accuracy is ``Static'', the prediction variance $\sigma_{i}$ of every $LFE_i$ is constant during the simulations. However, in the real world, the accuracy of flexibility predictions can vary depending on the weather, season, and type of DER. The ``{\em Dynamic}'' LFE accuracy configuration models a realistic setting in which the variance $\sigma_{i}$ of every $LFE_i$ fluctuates during the simulations. In particular, at each timeslot, the variance of the flexibility prediction accuracy of each LFE changes by +/- $0.001$, following a stochastic trajectory whose direction is generally downwards for good predictors or upwards for bad predictors.


\subsection{Experimental Results}

We now present 
our results in detail. The results are the averages of 30 simulations with different properties and 2000 timeslots each. The code for the aggregator framework and all experimental results will be publicly available online upon acceptance. 


\subsubsection{Scenario 1: Static LFE accuracy and CRPS Grid-to-Aggregator Payment}
\label{sec:staticLFEaccuracyCRPS}
\begin{figure}[ht]
  \centering
  \includegraphics[width=1\linewidth]{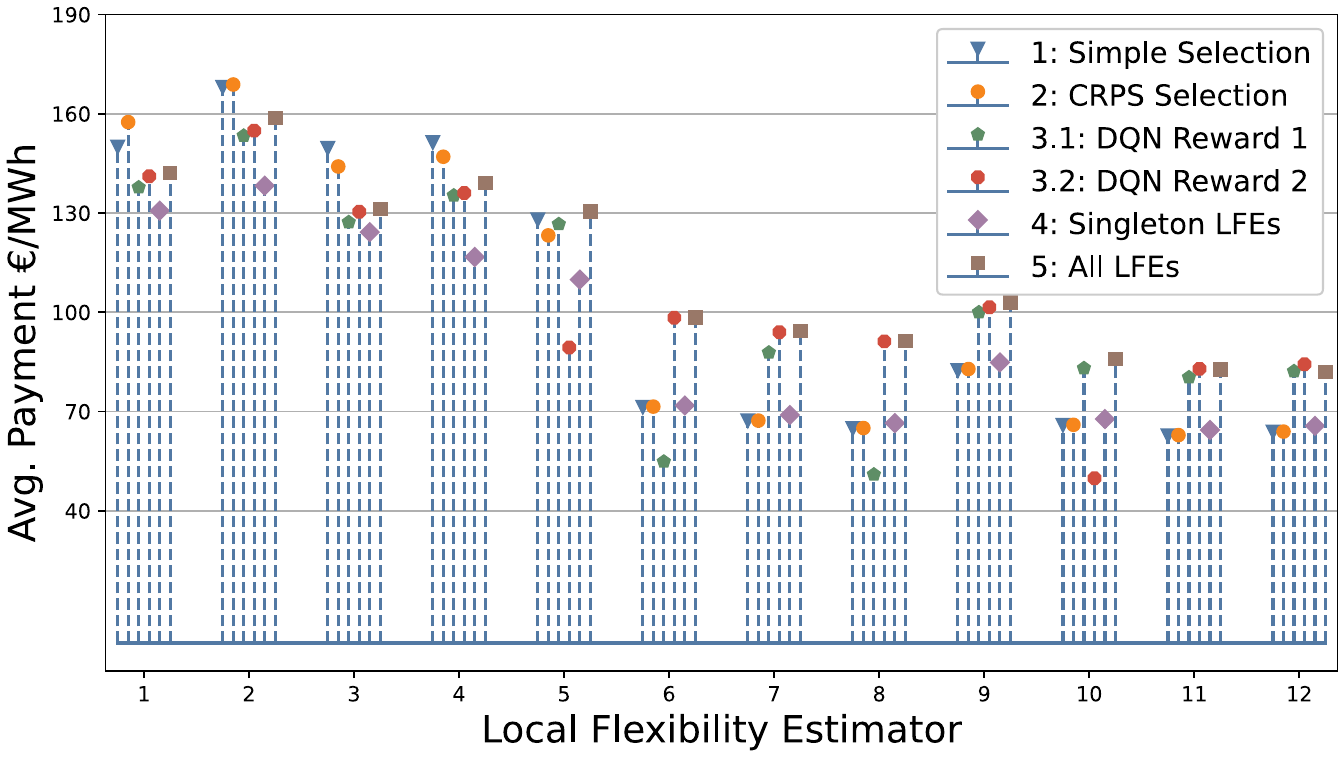}
  \caption{``Scenario 1'': Comparison of the average payment(€) per MWh sold of every LFE for all methods.}
  \label{fig:avg_static}
  \Description{``Scenario 1'': Comparison of the average payment(€) per MWh sold of every LFE for all methods.}
\end{figure}

The first scenario investigated in this work maintains the variance $\sigma_{i}$ of every $LFE_i$ constant during the simulations and utilizes the {\em CRPS Grid-to-Aggregator} payment (Eq.~\ref{eq:grid_agg_crps}). In Fig.~\ref{fig:avg_static}, $LFE_1$ resembles the best flexibility predictor with $\sigma_1 \approx 0$, progressing  gradually up to the worse predictor, $LFE_{12}$ with $\sigma_{12} \approx 1$.


Fig.~\ref{fig:avg_static} shows the average flexibility selling price for all the LFEs for different methods where LFEs' selection rules differ. At first, we can notice that the first four LFEs, which are the better predictors, are selling at a higher price than the rest, a result of both the prediction accuracy and the amount of flexibility they traded. The most rewarding methods for the ``best predictors'' are the {\em Simple Selection} or the {\em CRPS} ones. The most profitable methods for the ``worse predictors'' are when every LFE participates in the Aggregator without any selection mechanism; in this case, they are able to realize higher average payments due to the high total flexibility provided by the Aggregator as a whole, despite the fact that these payments are significantly lower than those of the better predictors (due to {\em CRPS} ``punishments'') even in this method. 

\begin{figure}[ht]
  \centering
  \includegraphics[width=1\linewidth]{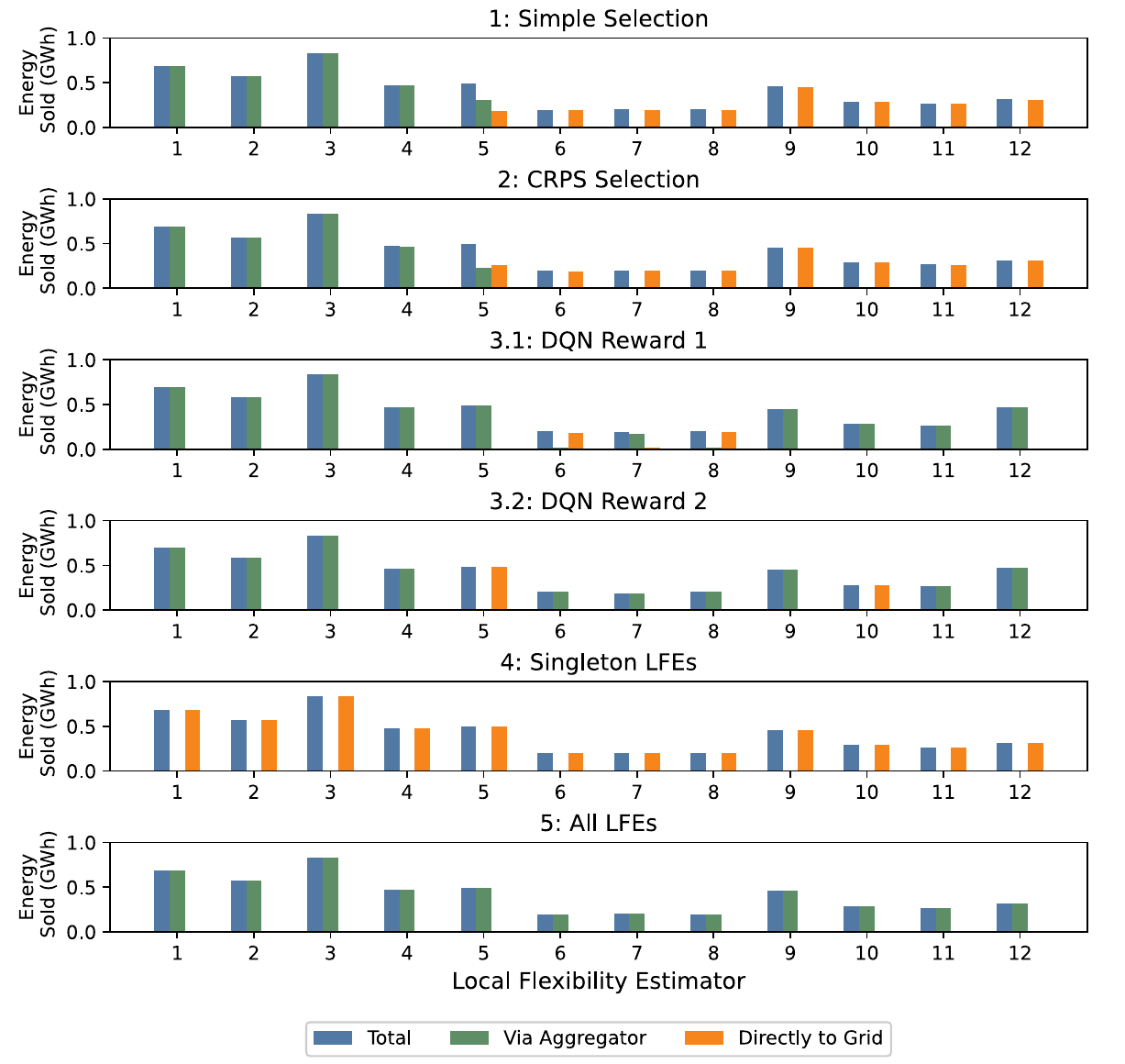}
  \caption{``Scenarios 1 \& 3'': Comparing the flexibility selling channel (directly to Grid or via the Aggregator).} 
  \label{fig:energy_sold_static}
  \Description{"Static LFE accuracy": Comparing the flexibility selling channel (directly to Grid or via the Aggregator) and the total flexibility (GWh) sold}
\end{figure}

To complete this analysis, it is necessary to understand how often the LFEs were selected to participate in the Aggregator, by observing their flexibility selling channel (Fig.~\ref{fig:energy_sold_static}). We can see that the {\em Simple} and {\em CRPS Selection} methods choose the most accurate LFE predictors ($LFE_{1,2,3,4,5}$) to participate and trade flexibility with the Aggregator; while the rest are not selected because of their low accuracy. In the {\em DQN Selection} methods, most LFEs are participating in the Aggregator, with {\em DQN} being able to distinguish somewhat better among medium and bad LFE predictors when the second reward function ($Reward_2$) is used (in which case the ``medium'' predictors do get selected to participate in Aggregator trade).


In summary, it is more profitable for the best LFEs to be chosen via the {\em Simple} and {\em CRPS Selection}, because these mechanisms reward truthful and reliable LFEs more. By contrast, less accurate LFEs receive higher payments when there is no selection mechanism in place (``{\em All LFEs participate in the Aggregator}'' method). 

\subsubsection{Scenario 2: Dynamic LFE accuracy and CRPS Grid-to-Aggrega-tor Payment}

\begin{figure}[ht]
  \centering
  \includegraphics[width=1\linewidth]{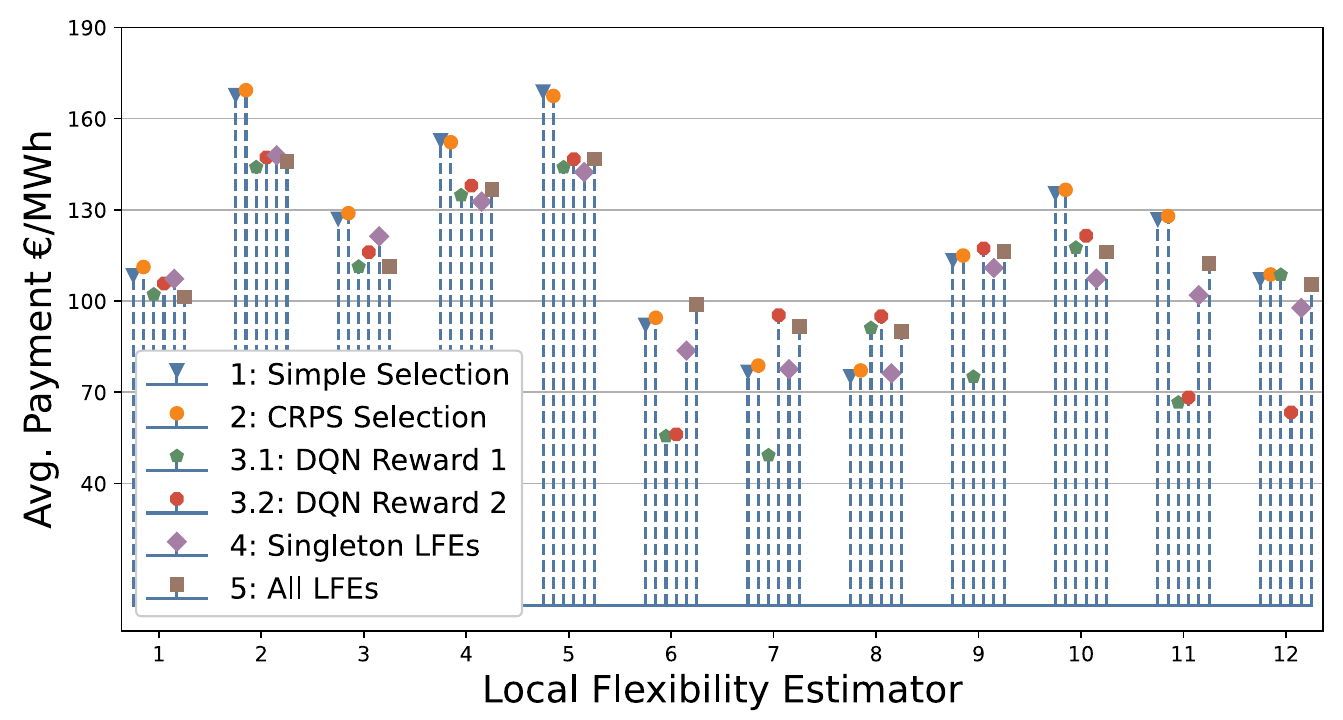}
  \caption{``Scenario 2'': Comparison of the average payment(€) per MWh sold of every LFE. }
  \label{fig:avg_dynamic}
  \Description{"Dynamic LFE accuracy": Comparison of the average payment(€) per MWh sold for every LFE}
\end{figure}

\begin{figure}[ht]
  \centering
  \includegraphics[width=1\linewidth]{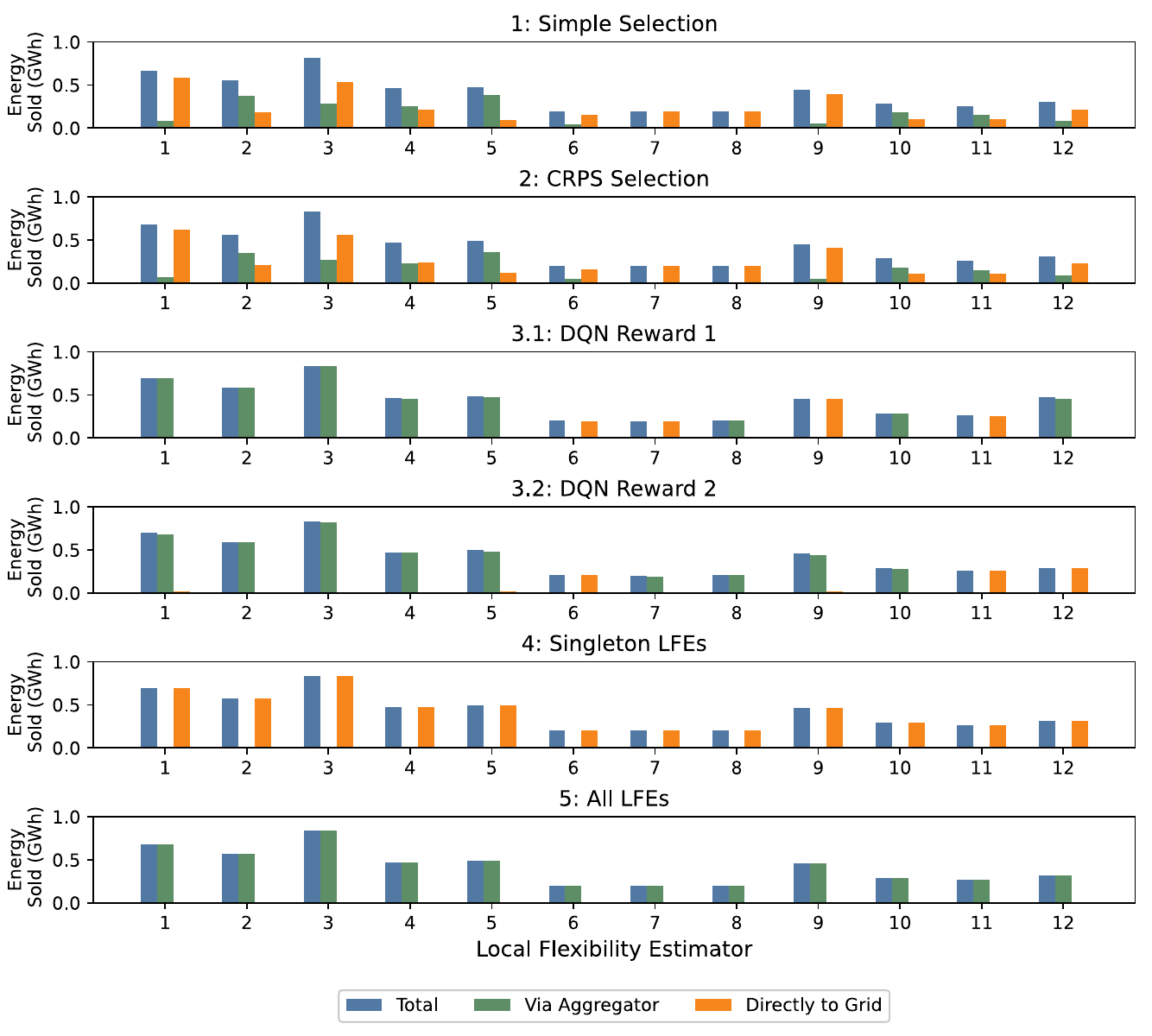}
  \caption{``Scenarios 2 \& 4'': Comparing the flexibility selling channel (directly to Grid or via the Aggregator). }
  \label{fig:energy_sold_dynamic}
  \Description{"Static LFE accuracy": Comparing the flexibility selling channel (directly to Grid or via the Aggregator)}
\end{figure}

The ``{\em Dynamic LFE accuracy}'' use case models a more realistic setting in which the ``worse predictor'' LFEs gradually improve while the best-performing ones degrade. 
It is obvious in Fig.~\ref{fig:avg_dynamic} that the payments attained by the LFEs are, in general, lower in this dynamic scenario. The {\em CRPS} and {\em Simple Selection} methods result in the best average payments for almost every LFE. Only the $LFEs_{6,7,8}$ had another method that was better than the first two selection mechanisms. In this use case, {\em DQN Selection} results in LFE payments that are, in many cases, comparable to those of the first two methods. However, this is not due to {\em DQN} being able to adapt to the dynamically changing capabilities of the agents.
Indeed, as seen in Fig.~\ref{fig:energy_sold_dynamic}, {\em DQN} generally selected the same LFEs to participate; while {\em Simple Selection} and {\em CRPS} are apparently able to detect the fluctuations in LFEs' prediction performance and now give much more participation chances to almost all LFE agents.


Finally, we report that for ``Scenario 1'' the Aggregator receives average payments (per MWh) of $151$€ using methods 1 and 2, while it only gets $120$€ per MWh in the rest. Similarly, for ``Scenario 2'' the aggregator gets an average payment of $168$€ per MWh using the first two methods and only $120$€ for the rest. These values highlight the fact that the Aggregator is able to make a higher profit per MWh when selecting the most accurate LFEs predictors via either {\em Simple} or {\em CRPS} selectors; and this is the case even in dynamic settings, where the total energy sold via the Aggregator is lower (specifically $\sim2$ GWh for both scenarios 1 \& 2) than in the static setting ($\sim3$ GWh). These values make intuitive sense since with method 5, all LFEs participate in the aggregator, and the {\em DQN} method selects most LFEs to participate in the Aggregator most of the time (cf. Fig.~\ref{fig:energy_sold_static} and Fig.~\ref{fig:energy_sold_dynamic}); while {\em Simple Scoring} and {\em CRPS} are more selective, thus the total amount of Aggregator flexibility (and thus its average payments) is higher.  Apart from these mechanisms' apparent positive effect on Aggregator efficiency, we note again that the use of selection mechanisms that incentivize reliable LFEs, thus promoting Grid stability, results in higher average payments for those LFEs (while all available flexibility is traded in any case). 

Overall, in both scenarios 1 and 2, we can see that it is not efficient for the LFEs to trade directly with the Grid, judging by the results of the ``{\em Singleton LFEs}'' method (Fig.~\ref{fig:avg_static}, Fig.~\ref{fig:avg_dynamic}). 
Furthermore, it is clear from our figures that {\em CRPS Selection} leads to higher payments for selected accurate LFEs. 
Also, the performance of the {\em Simple Selection} method is comparable to that of {\em CRPS}, both for the static and the dynamic settings. This, as explained in the introduction, marks an interesting trade-off between simplicity and guaranteed truthfulness of the selection mechanism to be used.


\subsubsection{Scenario 3: Static LFE accuracy and Simple Grid-to-Aggregator Payment}

We also conducted experiments simulating the way the existing aggregators are paid by the Grid when trading flexibility---i.e., using the {\em Simple Grid-to-Aggregator Payment} (Eq.~\ref{eq:simple_payment}).
\begin{figure}[!ht]
  \centering
  \includegraphics[width=1\linewidth]{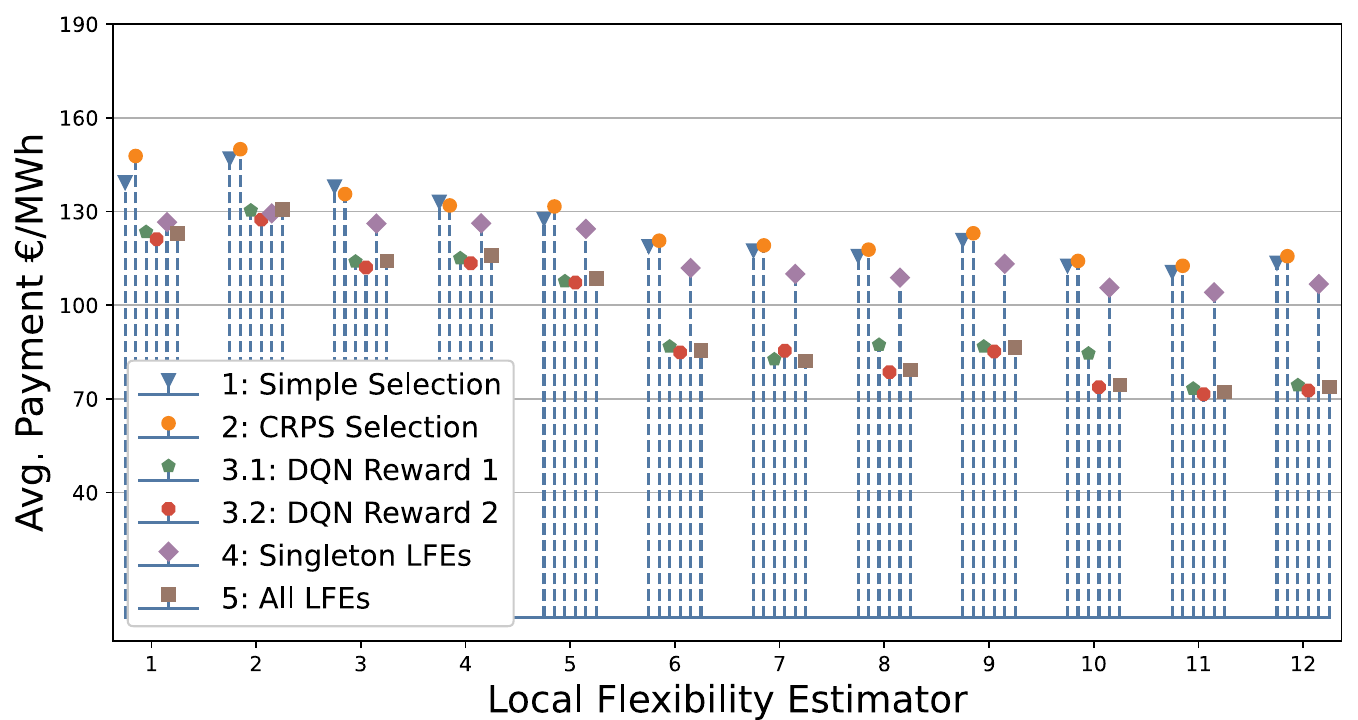}
  \caption{``Scenario 3'': Comparison of the average payment(€) per MWh sold of every LFE. }
  \label{fig:avg_static_std}
\end{figure}

Fig.~\ref{fig:avg_static_std} presents the average payment per MWh for every LFE.\footnote{Note that the average payments of Scenarios 1 and 2 are not comparable with the respective payments of Scenarios 3 and 4, since the Grid-to-Aggregator payments are completely different (Eq.~\ref{eq:grid_agg_crps} and \ref{eq:simple_payment}).} This figure verifies the previous findings, clearly depicting the superiority of the {\em CRPS Selection} method deployed by our framework. The {\em Simple Selection} method is the runner-up accumulating an average payment of a few more euros for $LFE_3$ and $LFE_4$, while in every other case, it scores lower than CRPS (albeit by a small margin). The ``{\em Singleton LFEs}'' baseline method is the third most efficient option, ranking, however, consistently and clearly lower than the first two. Notice that 
it can ``score'' $20$ euros less than {\em CRPS Selection}, a significant difference since each LFE sells many MWh. The remaining methods rank much lower, with a clearly poorer performance varying up to $30$ euros less than {\em CRPS Selection}.

The selection percentages of each LFE by the Aggregator agent remain the same with those of the``{\em Static LFE Accuracy}'' (Sec.~\ref{sec:staticLFEaccuracyCRPS}) configuration (i.e., are as in Fig.~\ref{fig:energy_sold_static}), because we maintained the same random seeds for the respective simulations. It was our intention to ensure that the offered flexibility of each LFE remains the same across different experiments so that we can fairly compare the rankings  (in terms of average payments) of the different selection methods across the various scenarios.

\subsubsection{Scenario 4: Dynamic LFE accuracy and Simple Grid-to-Aggre-gator Payment}

Our final experimental scenario is one with ``Dynamic'' LFE prediction accuracy and {\em Simple Grid-to-Aggregator Payments}. Here, the selection percentages of each LFE by the Aggregator are as in Fig.~\ref{fig:energy_sold_dynamic}).
{\em CRPS Selection} again consistently results to the highest average payment per MWh per LFE regardless of its prediction accuracy (bar LFEs 1 and 12 for which it is tied with {\em Singleton LFEs}). The {\em Singleton LFEs} method is a close runner-up, with payments that range from a few to 10 euros lower. The {\em Simple Selection} method is quite profitable too, rewarding LFEs on average with 5 euros less per MWh than {\em CRPS Selection}. On the other hand, the {\em DQN Selection} methods and, importantly, the {\em All LFEs} ``traditional'' Aggregator method (which essentially includes all available DERs in the Aggregator, as is current practice in the industry) result in average payments that are (at least) $30$ euros {\em lower} per MWh.

\begin{figure}[!ht]
  \centering
  \includegraphics[width=1\linewidth]{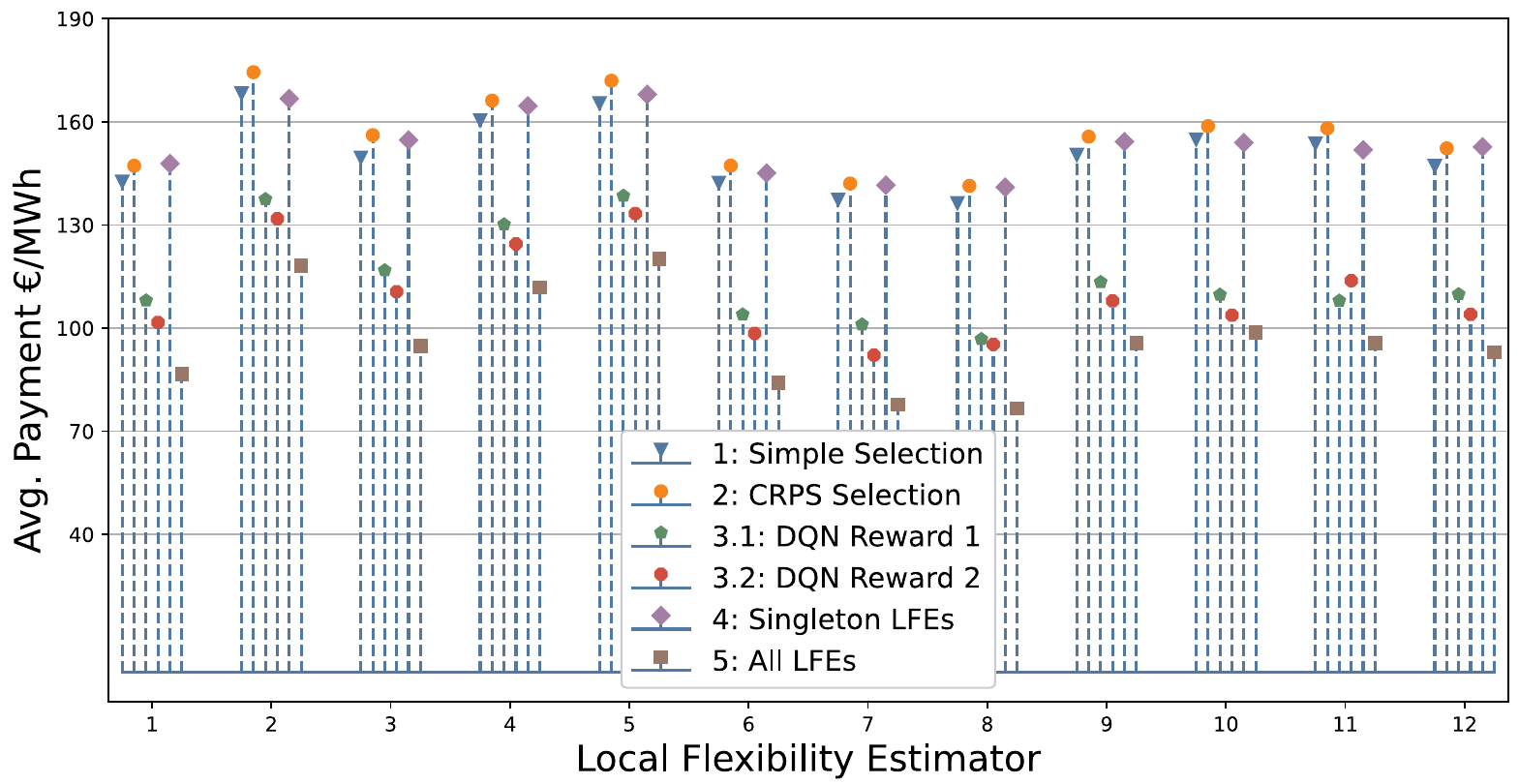}
  \caption{``Scenario 4'': Comparison of the average payment(€) per MWh sold of every LFE. }
  \label{fig:avg_dynamic_std}
\end{figure}

The findings of the last two scenarios, in which the Grid pays our Aggregator with a ``simple'' pricing mechanism that does not penalize flexibility prediction inaccuracies (Eq.~\ref{eq:simple_payment}) but only considers the total flexibility delivered, are quite interesting. 
They demonstrate that our aggregation framework, in particular when using the {\em CRPS Selection} method, {\em consistently achieves the highest average profit per LFE}, regardless of the LFEs’ prediction accuracy, and regardless of its dynamic or static nature (cf.~Fig.~\ref{fig:avg_static_std} and Fig.~\ref{fig:avg_dynamic_std}). Notice that the aforementioned ``simple'' pricing mechanism is, in fact, used in the current day-ahead electricity markets. Also, as mentioned, {\em CRPS Selection} is the clear winner when compared with the currently-in-use aggregation method that simply collects all available DERs' flexibility. Therefore, these results indicate that our framework using {\em CRPS Selection} can result in increased profits for LFEs (and subsequently DERs), if applied in the current Grid.

\section{Conclusions and future work}

In this work, we put forward a novel flexibility aggregation framework 
comprising a novel multiagent architecture along with various
(existing or novel) selection and pricing mechanisms. 
We conducted a systematic experimental evaluation, using data from the highly realistic PowerTAC simulator which we extended to allow for the incorporation of flexibility aggregators and related entities and mechanisms. 
Our results show that our framework and methods can successfully contribute to the effective integration of DERs in the Grid, enabling them to increase their profits.

In terms of future work, 
we intend to experiment 
with alternative selection and pricing mechanisms; and 
to study 
scenarios (readily supported by our framework) that allow LFEs to 
replace inefficient DER assets. 
Moreover, it would be interesting  to incorporate Aggregators that 
create more than one LFE cooperative, potentially using different mechanisms for each depending on its attributes. 
Finally, enhancing our framework with the ability to include multiple Aggregators competing for the representation of efficient LFEs, is also interesting future work.

\bibliographystyle{ACM-Reference-Format}
\bibliography{sample-base}

\appendix

\end{document}